\newif\ifdraft

\documentclass[submission,copyright,creativecommons,noderivs]{eptcs}

\drafttrue % interno

\ifdraft
\providecommand{\event}{GRAPHITE 2014} % Name of the event you are submitting to
\else
\fi

\newcommand{\ImgScale}{0.9}

\usepackage[english]{babel}
\usepackage{amssymb}
\usepackage{listings}
\usepackage{graphicx}
\usepackage{caption}
\usepackage{subcaption}

\usepackage{tikz}
\usetikzlibrary{arrows}
\usetikzlibrary{backgrounds}	% drawing the background after the foreground
\usetikzlibrary{automata,positioning}
\usetikzlibrary{decorations.pathmorphing} % To have curved arrows

\usepackage{varwidth} %Per avere i blocchi di una lunghezza massima

\usepackage{hyperref}

\usepackage[ruled,lined]{algorithm2e}

\usepackage{multicol}
\newcommand{\idt}{\makebox[5mm][l]{}}  %% indentation

\newcommand{\vars}[1]{\ensuremath{\mathit{Vars}(#1)}}

\newcommand{\abox}{\ensuremath{\mathit{ABox}}}

\newcommand{\tbox}{\ensuremath{\mathit{TBox}}}

\newcommand{\cls}[1]{\ensuremath{\mathsf{#1}}}   % classes
\newcommand{\rol}[1]{\ensuremath{\mathsf{#1}}}   % relations
\newcommand{\cons}[1]{\ensuremath{\mathsf{#1}}}  % constants
\newcommand{\funct}[1]{\ensuremath{\mathsf{funct}~#1}} % functionality
\newcommand{\inv}[1]{\ensuremath{#1^{\,-}}}  % inverse
\newcommand{\acname}[1]{\ensuremath{\mathsf{#1}}}
\newcommand{\action}[4]{\ensuremath{\acname{#1}[#2]}:~#3~$\rightsquigarrow$~#4}  % \action{Name}{Variables}{Condition}{Effect}
\newcommand{\acshort}[3]{\ensuremath{\acname{#1}}:~#2~$\rightsquigarrow$~#3}  % \acshort{Name}{Condition}{Effect}
\newcommand{\acdyn}[3]{#2~$\rightsquigarrow_\acname{#1}$~#3}  % \acdyn{Name}{Abox}{Abox'}
\newcommand{\concrete}[1]{\ensuremath{\mathit{concrete}}(#1)}  % \concrete{query,tbox}

\newcommand{\intdom}{\ensuremath{\Delta}}
\newcommand{\intfun}{\ensuremath{\mathcal{I}}}

\newcommand{\adom}{\ensuremath{\textsc{adom}}}
\newcommand{\alp}{\ensuremath{\textsc{alph}}}
\newcommand{\answ}{\ensuremath{\textsc{ans}}}

\newcommand{\dl}{\ensuremath{\textsc{dl}}}
\newcommand{\sj}{\ensuremath{\textsc{sj}}}

\newtheorem{example}{Example}

\newcommand{\asgn}{\ensuremath{:=}}

\title{
Backwards State-space Reduction\\
for Planning
in %Description Logic based 
Dynamic Knowledge Bases
}
\author{
Valerio Senni
\institute{IMT Institute for Advanced Studies, Lucca, Italy}
\email{valerio.senni@imtlucca.it}
\and
Michele Stawowy
\institute{IMT Institute for Advanced Studies, Lucca, Italy}
\email{michele.stawowy@imtlucca.it}
}
\def\titlerunning{Backward State-space Reduction for Planning}
\def\authorrunning{Senni V. \& Stawowy M.}

\begin{document}
\maketitle

\begin{abstract}
In this paper we address the problem of planning in rich domains, where knowledge representation is a key aspect for managing the complexity and size of the planning domain. We follow the approach of Description Logic (DL) based Dynamic Knowledge Bases, where a state of the world is represented concisely by a (possibly changing) ABox and a (fixed) TBox containing the axioms, and actions that allow to change the content of the ABox. The plan goal is given in terms of satisfaction of a DL query. 
In this paper we start from a traditional forward planning algorithm and we propose a much more efficient variant by combining backward and forward search. In particular, we propose a {\em Backward State-space Reduction} technique that consists in two phases: first, an {\em Abstract Planning Graph}~$\mathcal{P}$ is created by using the {\em Abstract Backward Planning Algorithm}~(ABP), then the abstract planning graph~$\mathcal{P}$ is instantiated into a corresponding planning graph~$P$ by using the {\em Forward Plan Instantiation Algorithm}~(FPI). The advantage is that in the preliminary ABP phase we produce a {\em symbolic} plan that is a pattern to direct the search of the concrete plan. This can be seen as a kind of informed search where the preliminary backward phase is useful to discover properties of the state-space that can be used to direct the subsequent forward phase.
We evaluate the effectiveness of our ABP+FPI algorithm in the reduction of the explored planning domain by comparing it to a standard forward planning algorithm and applying both of them to a concrete business case study.

%%% progressive instantiation 
%Due to the large size of the planing domain, planning is largely unpractical for real-world applications.
%Despite the implicit representation of the knowledge state allowed by DLs, planning is still largely unpractical for real-world applications.

%%% backward algorithms are considered to be more efficient

%Parole chiave:
%planning,
%description logic + swrl rules,
%planning graph minimization,
%forward and backward planning,
%abstract planning graph,
%abduction,
%actions.
\end{abstract}

% -----------------

\section{Introduction}
\label{sec:Introduction}

% INTRODUCTION AND PROBLEM STATING \\

In this paper we address the problem of planning in rich domains.
In particular we consider the setting where a large amount of data is structured and maintained
through a knowledge management system. We adopt a very well known and accepted formalism 
that is based on Description Logics and provides very efficient reasoning services, that is the DL-Lite framework~\cite{DBLP:journals/jar/CalvaneseGLLR07}.

The focus on planning comes from the industrial needs of knowledge representation and handling, 
as well as of work-flow management. 
Indeed, the large amount of data and high level of concurrency 
may pose a difficulty in handling resources efficiently and avoiding inconsistent behaviours.
Planning can be used as a way of dynamically 
producing consistent work-flows on the basis of a given goal to reach. 
Our notion of Dynamic Knowledge Base, that changes according to actions executed by agents in the system and
which we use to define our planning domain, is based on the DL-Lite framework.

Declarative work-flow management has been explored in~\cite{Montali2010a} and in~\cite{DBLP:books/sp/yawl2010/PesicSA10}.
The first, defines a formal framework combining the constraint-based language ConDec and Computational Logic,
where correctness of work-flow execution is granted by using verification techniques, run-time monitoring, and
post-execution consistency analysis. The second, again based on the language ConDec, allows to specify work-flows 
in a declarative way by using LTL-based temporal constraints. Both of these approaches do not explicitly deal
with data through an ontology as we can do using Description Logics.
Another approach to the construction of correct work-flows can be found in~\cite{Sirin2006},
where Hierarchical Task Network planning is applied to a Description Logic (DL) based planning domain.
This work had a different application domain with respect to ours, that is web service composition.
The idea of using DLs to represent the states of the world is also used in~\cite{DBLP:journals/jair/HaririCMGMF13},
where the notion of~\emph{Knowledge and Action Base} (KAB) is introduced.
Actions are applicable whenever a query has an answer in the current state of the world and has the effect of creating a new state.
This last work is the one closest to ours in spirit, because it targets the business domain. However, there are several differences with
our setting. First, the notion of planning domain. Second, the fact that they focus on the analysis of temporal properties 
of these evolving knowledge bases while we focus on the problem of state-space explosion when looking for a plan.

The contribution of this paper is the proposal of a novel planning technique that combines backwards
and forwards analysis to reduce the explored state-space. This technique exploits a symbolic representation
of states and reasoning techniques provided by Description Logics.

Forward search algorithms can be made very effective if an efficient search heuristics is provided.
However, in lack of such heuristics they may end up exploring several paths that are not useful for reaching the goal.
Backward search algorithms can take advantage of knowing the goal state to consider only actions that can lead to such a state.
Our planning technique has the purpose of reducing the explored state-space by combining backward and forward planning to take
advantage of both approaches. In particular, we propose a {\em Backward State-space Reduction} technique that consists in
structuring the planning in two phases (identified jointly as ABP+FPI): first, an {\em Abstract Planning Graph}~$\mathcal{P}$
is created by using the {\em Abstract Backward Planning Algorithm} (ABP), then the abstract planning graph~$\mathcal{P}$
is instantiated into a corresponding planning graph $P$ by using the {\em Forward Plan Instantiation Algorithm}~(FPI).

The advantage of the Backward State-space Reduction technique is that, in the preliminary ABP
phase, we produce (through a backward search) a {\em symbolic} plan that is a pattern to direct the search of the concrete plan.
In particular, in the subsequent (forward) FPI construction, the abstract plan constraints the {\em choice of actions} to be applied and
the actions have also {\em stronger application conditions}. This can be seen as a kind of informed search where the preliminary backward phase
is useful to discover properties of the state-space that can be used to direct the subsequent forward phase.
%In particular, the abstract planning graph~$\mathcal{P}$ can be understood as a set of constraints over actions applied in the
%forward construction of~$P$. 
The effect of constraints induced by the abstract plan is to significantly reduce the branching
of the planning by exploiting information propagated from the goal condition in a backward manner.

%Our effort in the development of the Backward State Space Reduction
%technique has been in making it as disjoint as possible from the actual reasoning mechanism of
%the underlying logical representation of knowledge. 

We have implemented the ABP+FPI algorithm and compared it to a standard Forward Planning algorithm over
a case study that was designed to scale according to the values of certain parameters.
The preliminary experimental results are promising both in terms of the time taken for finding 
the entire set of plans and in terms of the actual number of explored states.

First, we start by introducing the formalization of Dynamic Knowledge Base in Sec.~\ref{sec:Description Logic based dynamic systems}.
Then, we illustrate in Sec.~\ref{sec:Case Study} a Case Study which we use to test our planning technique.
In Sec.~\ref{sec:Planning in DLDSs} we describe the Backward State-space Reduction technique. 
Finally, in Sec.~\ref{sec:Experiments} we discuss a software implementation of our algorithm and some preliminary experiments.

% -----------------

%\newpage
\section{Dynamic Knowledge Bases}
\label{sec:Description Logic based dynamic systems}

For modelling domain resources and their relationships
we adopt the Description Logic (DL) framework~\cite{DBLP:conf/dlog/2003handbook}, which it is tailored to modelling 
a data domain by means of {\em concepts}, that are sets of individuals, and
{\em roles}, that are binary relations among individuals. 
%A DL Knowledge Base (KB) allows to represent data 
%{\em intentionally}, by expressing axioms over concepts and roles that must hold over all individuals of the domain, and
%{\em extensionally}, by explicitly mentioning individuals that belong to a concept and their participation into roles.
%Therefore, 
A DL knowledge base is made of two elements: a $\tbox$~$T$, containing axioms over concepts and roles that must hold over 
all individuals of the domain, and an $\abox$~$A$, containing membership assertions of individuals to concepts and
their participation into roles. Given a knowledge base $\langle T,A\rangle$, there is a number of reasoning
tasks that can be performed, among these the ones we are interested in are querying and consistency check.
%We recall in the following the notions that are essential for this paper.

Let us introduce the syntax of a fragment of DL, called DL-Lite~\cite{DBLP:journals/jar/CalvaneseGLLR07}.
Let~$\cls{N}$ be an atomic concept name and~$\rol{P}$ be an atomic role name. We can compose them by using constructors in order
to define more complex concepts and roles as given in the following grammar:\\[1mm]
\makebox[10mm][l]{}\makebox[30mm][l]{$B~~:=~~\cls{N}~\mid~\exists~R$}\makebox[30mm][l]{$C~~:=~~B~\mid~\neg~B$} (composed concepts)\\[1mm]
\makebox[10mm][l]{}\makebox[30mm][l]{$R~~:=~~\rol{P}~\mid~\inv{\rol{P}}$  }\makebox[30mm][l]{$V~~:=~~R~\mid~\neg~R$} (composed roles)\\[1mm]
where~$\exists$ is the projection of~$R$ over its first argument, $\inv{}$ is the inverse relation, and
$\neg$~is the complement. \tbox~axioms are constructed from (composed) concepts and roles according to the following schemes:\\[1mm]
\makebox[10mm][l]{}\makebox[50mm][l]{$B \sqsubseteq C$~~~(concept inclusion)}\makebox[50mm][l]{$R \sqsubseteq V$~~~(role inclusion)}\makebox[50mm][l]{$\funct{R}$~~~(functionality)}\\[1mm]
They can be translated into equivalent FOL formulas in a standard way.
The \abox{} contains ground instances of concepts and roles, such as $\cls{N}(\cons{a})$ and $\rol{P}(\cons{b},\cons{c})$, for some
individuals represented by the constants~\cons{a}, \cons{b}, \cons{c}.
For the sake of simplicity, in this paper we do not allow function symbols, thus we consider only finitely
many possible individuals in a knowledge base. Extensions of the DL-Lite framework allowing to reason over equalities
and possibly infinite sets of individuals constructed from a finite signature are possible~\cite{DBLP:journals/jair/ArtaleCKZ09}.
We plan to extend the techniques presented in this work to that more general setting.
\begin{example}\label{ex:kb-simple}
We can model the hierarchy of employees within a company as follows.\\[1mm]
\idt$T~=~\{\cls{Technician} \sqsubseteq \cls{Employee},~\cls{Manager} \sqsubseteq \cls{Employee},~\cls{Technician} \sqsubseteq \neg\, \cls{Manager}\}$\\[1mm]
Where the~$\tbox$ requires that technicians and managers are employees, and they are disjoint concepts (no individual can be
part of both). One possible $\abox$ is the following:\\[1mm]
\idt$A~=~\{\cls{Technician}(\cons{e002})\}$\\[1mm]
where an individual identified by the constant~$\cons{e002}$ is classified as a~$\cls{Technician}$.
The assignment of responsibilities of managing documents is modeled by a role $\rol{assignedTo}$ and further axioms as follows:\\[1mm]
\idt$T'~=~T\cup\{\exists~\rol{assignedTo} \sqsubseteq \cls{Document},~\exists~\inv{\rol{assignedTo}} \sqsubseteq \cls{Employee},~\funct{\rol{assignedTo}}\}$\\[1mm]
Note that we set (i)~the domain of the role~$\rol{assignedTo}$ to be included within~$\cls{Document}$
(using projection), (ii)~the range of the role~$\rol{assignedTo}$ to be included within~$\cls{Employee}$ (using projection and the inverse operator),
and (iii)~$\rol{assignedTo}$ to be functional (no document can be assigned to two employees at the same time).
We will discuss an extension of this example in the case study in Sec.~\ref{sec:Case Study}.
\end{example}

We now introduce reasoning tasks over DL-Lite knowledge bases that we use in the planning algorithm:
testing consistency (i.e. the existence of a model) and answering queries.
A DL-Lite knowledge base is interpreted following the standard First Order Logic approach, where we fix
a domain of interpretation~$\intdom$ and an interpretation function~$\intfun$ mapping individuals to elements of~$\intdom$,
concepts to subsets of~$\intdom$, and roles to subsets of~$\intdom\times\intdom$. DL-Lite axioms can be translated into
equivalent FOL formulas and interpreted accordingly.
An interpretation~$\intfun$ is a {\em model} of a knowledge base~$\langle T,A\rangle$ if it satisfies (the translation of)
all the assertions in~$T$ and~$A$.
A knowledge base is {\em satisfiable} if it has a model. Finally, an~$\abox$~$A$ is said to be {\em consistent}
w.r.t.~a~$\tbox$~$T$ if the knowledge base~$\langle T,A\rangle$ is satisfiable. 
\begin{example}\label{ex:kb-model}
One possible model for the knowledge base~$\langle T,A\rangle$ presented in Example~\ref{ex:kb-simple} 
is the one where $\cls{Technician}(\cons{e002})$ and $\cls{Employee}(\cons{e002})$ hold (obviously, considering only~$\cls{Technician}(\cons{e002})$
would not satisfy all the axioms). An example of an inconsistent $\abox$ is
$A'=\{\cls{Technician}(\cons{e002}),\cls{Manager}(\cons{e002})\}$, which contradicts the last axiom of~$T$.
\end{example}

Queries over a DL-Lite knowledge base~$\langle T,A\rangle$ are constructed considering (i)~the set of the constants appearing
in the assertions of~$A$, called the {\em active domain}~$\adom(A)$, and (ii)~the set of the predicate symbols occurring
in~$T$ and~$A$, called the {\em alphabet}~$\alp(T,A)$. A query~$q$ is a FOL formula of the
form~$\bigvee_{i=1}^n(\exists\,y_i.\,\mathit{conj}_i(\vec{x}_i,\vec{y}_i))$ called {\em union of conjunctive queries},
where~$\mathit{conj}_i(\vec{x}_i,\vec{y}_i)$ is a finite conjunction of atoms of the form~$\cls{N}(z)$ and~$\rol{P}(z,z')$,
for $\cls{N},\rol{P}\in\alp(T,A)$,
and~$z,z'$ are either variables in~$x_i\cup y_i$ or constants in~$\adom(A)$. The {\em certain answers} to a query~$q$
over~$\langle T,A\rangle$ is the set~$\answ(q,T,A)$ of substitutions~$\vartheta$ mapping free variables of~$q$
to constants in~$\adom(A)$ and such that~$q\vartheta$ holds in {\em every model} of~$\langle T,A\rangle$, 
that is,~$q\vartheta$ is a logical consequence of~$\langle T,A\rangle$.

The most interesting feature of the DL-Lite framework is computing the set of the certain answers to a query is decidable
and also efficient~\cite{DBLP:journals/jair/ArtaleCKZ09}, being PT\textsc{ime}-complete in the size of the~$\abox$ and the~$\tbox$
and~AC$^0$ in the size of the $\abox$ (this complexity is often referred to as the {\em data complexity}).

In some cases, it can be useful to explicitly specify a role in terms of the product of two concepts, rather than simply
constraining its domain and range, as discussed in the following Example~\ref{ex:product}.
\begin{example}\label{ex:product}
Let us consider the model of Example~\ref{ex:kb-simple} and, in particular, the knowledge base~$\langle T',A\rangle$ defined therein.
We can further categorize documents by introducing the subclass of technical documents through the axiom~$\cls{TechnicalDoc} \sqsubseteq \cls{Document}$.
It can be useful to introduce a $\rol{canManage}$ role, which expresses the competences of certain employees in managing
certain documents: e.g.~one can be interested in modelling that {\em every technician can manage every technical document}.
This can be used, for example, as a prerequisite for assigning documents to employees (e.g. technical documents cannot
be assigned to non-technicians). However, this cannot be expressed directly as a DL-Lite axiom, since the required axiom 
is~$\forall~x,y.~(\cls{Technician}(x)\wedge\cls{TechnicalDoc}(y) \rightarrow \rol{canManage}(x,y))$,
which falls outside the allowed syntax. What we can model in DL-Lite is given by the following axioms:
$\exists~\inv{\rol{canManage}} \sqsubseteq \cls{TechnicalDoc}$ and $\exists~\rol{canManage} \sqsubseteq \cls{Technician}$,
allowing only to constrain domain and range of the role of~$\rol{canManage}$.
\end{example}
In general neither joins nor concept products are expressible in DL-Lite and therefore we are not allowed to consider
axioms of the form~$\forall~x,y.~(\cls{N}_1(x)\wedge\cls{N}_2(y) \rightarrow \rol{R}(x,y))$, 
as mentioned in the example. In this paper we allow {\em this specific form}
of axioms on the basis of recent results obtained in a language called Datalog$^\pm$~\cite{DBLP:journals/ws/CaliGL12}.
Datalog$^\pm$ is indeed a family of languages, that strictly generalizes the DL-Lite family. The so-called {\em sticky} fragment 
of Datalog$^\pm$~\cite{DBLP:conf/rr/CaliGP10} allows to specify concept products, as well as other more general forms of joins,
and enjoys the same data complexity of DL-Lite. In general, DL-Lite axioms can be translated to Datalog$^\pm$ clauses.
However, for the purpose of this paper, we stick to the restricted case of the DL-Lite syntax and we simply allow the use of axioms
of the form~$\forall~x,y.~(\cls{N}_1(x)\wedge\cls{N}_2(y) \rightarrow \rol{R}(x,y))$ (or simply $\cls{N}_1(x)\wedge\cls{N}_2(y) \rightarrow \rol{R}(x,y)$)
in the \tbox, keeping the good complexity results. We call these axioms {\em simple joins} and we assume to identify the
DL-lite axioms in~$T$ by~$\dl(T)$ and the simple join axioms in~$T$ by~$\sj(T)$. We will use this separation
for the design of the planning algorithms. We call~$\rol{R}(x,y)$ the {\em conclusion} of the axiom and~$\cls{N}_1(x)\wedge\cls{N}_2(y)$
the {\em premise}. We leave for future work the extension of
our planning domain reduction technique to the full Datalog$^\pm$ framework.

Let us now consider the {\em dynamical} aspect of our knowledge bases. We introduce the notion of {\em Dynamic Knowledge Base} (DKB),
which is a transition system where states are DL-Lite knowledge bases and actions are used to update the \abox{}. In particular,
we assume the $\tbox$~$T$ does not change, so the ABox $A$ is sufficient to identify the state of the system and we will feel free
to refer only to $A$, without explicitly mentioning $T$. A Dynamic Knowledge Base is a tuple $\langle T,A_0,\Gamma\rangle$,
where $T$ is a TBox, $A_0$ is an ABox called the {\em initial state}, and $\Gamma$ is a finite set of well-formed actions.

An {\em action} is of the form~\action{a}{x_1,\ldots,x_n}{$q$}{$e$} (or simply \acshort{a}{$q$}{$e$}),
where $\acname{a}$ is the action name, $q$ is a query (called {\em action guard}),
and $e$ is an (possibly non-ground) $\abox$ assertion (called {\em action effect}) 
such that $\vars{e}\subseteq\vars{q}=\{x_1,\ldots,x_n\}$.
An action is {\em well formed} w.r.t.~a knowledge base $\langle T,A\rangle$ if predicates occurring in its effects are disjoint
from predicates occurring in conclusions of simple join axioms in~$\sj(T)$.
The well-formedness of actions is an important condition since it allows us to distinguish the assertions that are entailed by simple join axioms from the assertions that can be introduced by actions.
The informal semantics of an action is that, by using the query~$q$, we extract {\em one} certain answer~$\vartheta$ from the current knowledge base and we obtain a corresponding {\em ground} $\abox$ assertion~$e\vartheta$.
The effect of an action~$\acname{a}$ over a state~$A$ is, non-deterministically, a new state~$A\cup \{e\vartheta\}$, for each~$\vartheta\in\answ(q,T,A)$, which we indicate as \acdyn{a,\vartheta}{$A$}{$A\cup\{e\vartheta\}$}.
We call \acname{a}$\vartheta$ an {\em instantiation} of \acname{a}. 

Of course adding an assertion to an $\abox$ could make it inconsistent.
Such cases are not considered in our transition system, and actions' instantiations that lead to inconsistent \abox-es are ignored.

\begin{example}\label{ex:action}
Following up our previous examples based on a company case study, we can model the fact that a manager can decide to assign
documents to employees, e.g.~for revision purposes. The following action requires to identify a manager and an employee that
is able to manage a document, and under these conditions the document can be assigned to that employee:\\[1mm]
\idt\action{appoint}{x,y,z}{$\cls{Manager}(x)\wedge \rol{canManage}(y,z)$}{$\rol{assignedTo}(z,y)$}\\[1mm]
Considering Example~\ref{ex:product}, if~$y$ is bound to a technical document, the only eligible employees will be technicians.
The effect on the knowledge base is the addition of an $\rol{assignedTo}$ assertion. Now, assume an $\abox$ of the form 
$A=\{\cls{Manager}(\cons{e001}),\cls{Technician}(\cons{e002}),\cls{Technician}(\cons{e003}),\cls{TechnicalDoc}(\cons{d001})\}$, 
where we have two technicians. The two possible effects of the action $\acname{appoint}$ are\\[1mm]
\makebox[6mm][l]{}\acdyn{appoint,\vartheta_1}{$A$}{$A\cup\{\rol{assignedTo}(\cons{d001},\cons{e002})\}$}~~~and~~~
\acdyn{appoint,\vartheta_2}{$A$}{$A\cup\{\rol{assignedTo}(\cons{d001},\cons{e003})\}$},\\[1mm]
where~$\vartheta_1 =\{x\mapsto\cons{e001},y\mapsto\cons{e002},z\mapsto\cons{d001}\}$ 
and~$\vartheta_2 =\{x\mapsto\cons{e001},y\mapsto\cons{e003},z\mapsto\cons{d001}\}$. \\
As mentioned before, actions can lead to inconsistent states:\\[1mm]
\makebox[6mm][l]{}\acdyn{appoint,\vartheta_3}{$A\cup\{\rol{assignedTo}(\cons{d001},\cons{e002})\}$}{$A\cup\{\rol{assignedTo}(\cons{d001},\cons{e002}),\cls{assignedTo}(\cons{d001},\cons{e003})\}$},\\[1mm]
where the axiom regarding the functionality of the role $\rol{assignedTo}$ is violated.
Such transition would be discarded in the transition system.
\end{example}

The notion of Knowledge and Action Base presented in~\cite{DBLP:journals/jair/HaririCMGMF13} is in principle very similar to our DKB and has been proposed to model knowledge base dynamics.
Indeed, we have been inspired by that notion, but the main difference between our definition and that one is in the way the new knowledge base is constructed as an effect of an action.
In particular, in their approach, no assertion from the previous $\abox$ is maintained and the new $\abox$ is entirely constructed summing the effect of all possible instantiations~$\acname{a}\vartheta$ of an action~$\acname{a}$.
Furthermore, actions have also a more general form with various possible effects applied in {\em parallel} and {\em at once}.
Therefore, that definition implements a semantics of the dynamics which is very different form ours.
In their setting, however, our semantics can be reconstructed by explicitly adding actions that reconstruct the information that we preserve at each step.
Discussions on the appropriateness of these models are longstanding and related to the frame problem as well as to other well known problems in the field of knowledge representation~\cite{reiter2001knowledge}.
In this paper we do not address these problems because we focus on planning and state-space reduction aspects.
We leave for future work to study how our state-space reduction technique adapt to different
dynamic models.

%%% \vcomm{typical problems of reasoning with actions: frame problem, ramification problem, qualification problem [Reiter 2001]}
%The set of {\em certain answers} to a query $q$ in a KB $\langle T,A\rangle$ is defined as
%$\ans{q,T,A}=\{\vartheta\ \mid\ T,A,\vartheta\models q\}$
%%\newpage
%Recent work on DL-based dynamical systems, 1 page:
%\begin{itemize}
%\item motivations and comparison w recent work on the topic (De Giacomo ...)
%\item definition of state transition system (sate = Abox+Tbox, only Abox is changing, and transitions by add/remove actions)
%\end{itemize}

% -----------------

%\newpage
\section{Case Study}
\label{sec:Case Study}

%Case Study (1 page)
%
%This part is giving the motivation to the paper and the exemplification of the various aspects (what is a state, how do we change state).
%In this section we present the case study and we give some motivations for it. We state the goal here.
%Later on, in other sections, we will consider again the case study and we show how by various techniques (fw/bw) we solve the goal.
%
%General description of the case study
%We want to be able to model:
%- classes of distinguished elements
%- hierarchy
%- roles between classes, classes that can participate in a role only with specific classes
%- agents' actions

In our case study a company is interested into having a centralized control over its activities, with the aim of minimizing 
the effort of maintaining the consistency of the entire set of its work-flows.
In the company, employees are organized in specific areas of competence and documents are assigned accordingly.
In this setting, we imagine that the work of an employee is assisted by a planning software,
which receives as input an high level goal and provides one or more plan to reach such goal.
The planner should satisfy the company ruled represented by axioms in the $\tbox$.

\begin{figure}[t]
\vspace{-4mm}
\centering
\begin{tikzpicture}[->,auto,>=latex', node distance=1.5cm, on grid, font=\small, scale=\ImgScale, every node/.style={transform shape}] 
\tikzstyle{state}=[draw]
	
\node[state] (employee) {$\cls{Employee}$}; 
\node[state] (technician) [below right of=employee] {$\cls{Technician}$};
\node[state] (administrative) [below of=employee, node distance=2cm] {$\cls{Administrative}$};
\node[state] (manager) [below left=of employee] {$\cls{Manager}$};

\path 
(employee) edge[] node {} (technician) 
(employee) edge[] node {} (administrative) 
(employee) edge[] node {} (manager) 
; 

\node[state] (document) [right of=employee, node distance=6cm] {$\cls{Document}$}; 
\node[state] (technical) [below left of=document] {$\cls{TechDoc}$};
\node[state] (urgent) [below right of=document] {$\cls{UrgentDoc}$};
\node[state] (admin) [below of=document, node distance=2cm] {$\cls{AdmDoc}$};

\path 
(document) edge[] node {} (technical) 
(document) edge[] node {} (admin) 
(document) edge[] node {} (urgent) 

(document) edge[above, dashed] node {$\funct{\rol{assignedTo}}$} (employee) 
(technician) edge[above, dashed] node {$\rol{canManage}$} (technical) 
(administrative) edge[above, dashed] node {$\rol{canManage}$} (admin) 
; 

\node[state] (docState) [right of=document, node distance=4cm] {$\cls{DocumentState}$}; 

\path 
(document) edge[above, dashed] node {$\rol{hasStatus}$} (docState) 
; 
\end{tikzpicture}
\caption{Graphical representation of the $\tbox$} \label{fig:case study - tbox}
\vspace{-4mm}
\end{figure}
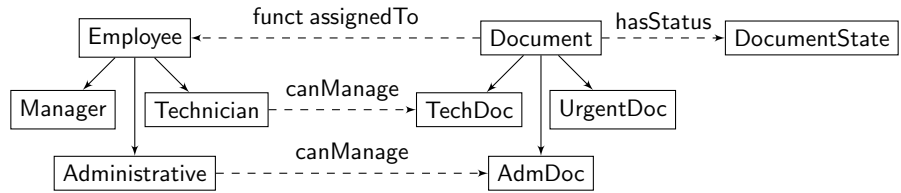

In Fig.~\ref{fig:case study - tbox} we give an informal representation of the hierarchy of employees and of documents
in the company (identified by solid boxes and connected through solid lines). Roles involving these concepts of individuals 
are represented through dashed lines.
This structure has been partially formalized by axioms discussed in examples given in Sec.~\ref{sec:Description Logic based dynamic systems}.
Now we illustrate more extensively how this structure is formalized by using DL-Lite and Simple Join axioms.
The complete specification of the case study can be found in the \nameref{sec:Appendix}.

The agents of the company are grouped in the concept $\cls{Employee}$. 
As explained in Example \ref{ex:kb-simple}, we build a hierarchy by introducing the concepts 
$\cls{Manager}$, $\cls{Administrative}$ and $\cls{Technician}$ and using axioms of the 
type~$\cls{Technician} \sqsubseteq \cls{Employee}$ to denote concept containment.
Disjointness between concepts is specified using axioms like $\cls{Technician} \sqsubseteq \neg \cls{Manager}$.
Similarly, we categorize the documents within the main concept $\cls{Document}$, by introducing 
$\cls{TechnicalDoc}$, $\cls{AdministrativeDoc}$, and $\cls{UrgentDoc}$, where $\cls{TechnicalDoc}$ is disjoint from $\cls{AdministrativeDoc}$.
We also create a concept $\cls{DocumentState}$ which expresses the status of a document, for example $\cons{reviewed}$.

%Instances are placed inside concepts using $\abox$ assertions like $\cls{Technician}(\cons{e002})$.
For the case study we consider different numbers of employees and documents by considering different sets of 
assertions like $\cls{Technician}(\cons{e002})$ in the \abox.
We discuss this aspect more in details in Sec.~\ref{sec:Experiments}, planning experiments and performance of the algorithms
presented in Sec.~\ref{sec:Planning in DLDSs} are discussed.

We define also roles in which instances can participate. The role $\rol{hasStatus}$ relates 
a document to a working status (like $\cons{reviewed}$).
The roles ($\rol{canManage}$ and $\rol{assignedTo}$) describe the relationship between the employees and 
documents, allowing us to put some restrictions on the type of documents an employee can deal with.
In particular, $\rol{canManage}$ specifies which employees can manage which documents,
while $\rol{assignedTo}$ specifies which documents have been assigned to a specific employee.

The axiom~$\cls{Technician} \sqsubseteq \exists \rol{canManage}.\cls{TechnicalDoc}$ we require that the concept \cls{Technician} is a subconcept
of the domain of $\rol{canManage}$, but restricted only to the assertions in which the range are instances from the concept $\cls{TechnicalDoc}$.
Consider for example the assertion $\rol{canManage}(\cons{e004}, \cons{d001})$: if $\cons{d001}$ is not a technical document, then $\cons{e004}$ 
cannot be a technician. We define a similar restriction over the concept $\cls{Administrative}$, by using the axiom~$\cls{Administrative} \sqsubseteq \exists \rol{canManage}.\cls{AdministrativeDoc}$.

On top of the simple join axiom defined in Example \ref{ex:product}, we also introduce an axiom relative to $\cls{Administrative}$ employees
 and corresponding $\cls{AdministrativeDoc}$: \\[1mm]
\idt $\forall~x,y.~(\cls{Administrative}(x) \wedge \cls{AdministrativeDoc}(y) \rightarrow \rol{canManage}(x,y))$ \\[1mm]
which defines that every pair of administrative employee and document is in the role $\rol{canManage}$.
The actions available in the set $\Gamma$ are $\acname{appoint}$ (defined Example \ref{ex:action}), $\acname{review}$, $\acname{setAdmDoc}$, and $\acname{setTechnician}$.
The action $\acname{review}$ allows to set a document to the $\cons{reviewed}$ state if it is in a relationship $\rol{assignedTo}$.
\idt\action{review}{x,y}{$\rol{assignedTo}(x,y)$}{$\rol{hasStatus}(x,\cons{reviewed})$} \\
A manager can set a document (resp., an employee) as an administrative document (resp., a technician) through the action $\acname{setAdmDoc}$ (resp., $\acname{setTechnician}$) given below: \\
\idt\action{setAdmDoc}{x,y}{$\cls{Manager}(x) \wedge \cls{Document}(y)$}{$\cls{AdministrativeDoc}(y)$} \\
\idt\action{setTechnician}{x,y}{$\cls{Manager}(x) \wedge \cls{Employee}(y)$}{$\cls{Technician}(y)$}

In this case study, the objective is to reach a state where an urgent document reaches the revised state,
which is expressed in terms of a {\em goal} as follows:~$\cls{UrgentDoc}(x)\wedge\rol{hasStatus}(x, \cons{reviewed})$.
This goal is provided to the planner for the construction of a plan that allows us to realize it.

% -----------------

%\newpage
\section{Planning in Dynamic Knowledge Bases}
\label{sec:Planning in DLDSs}

We consider the classical reachability planning problem~\cite{Nau:2004:APT:975615}
in the domain of dynamic knowledge bases and forward and backward search algorithms.
We assume {\em uninformed} algorithms~\cite{Russell2010}, as we want them to be usable in different scenarios, where there could
be no obvious heuristic function to guide the algorithm. %, mainly because in dynamic knowledge bases the states are very rich structures. 
However, our algorithms can be easily parametrized to add an appropriate search heuristic.

Given a dynamic knowledge base $\langle T, A_0, \Gamma \rangle$, formalized in Sec.~\ref{sec:Description Logic based dynamic systems},
we define our \emph{planning problem} as the tuple $\langle T, A_0, \Gamma, g \rangle$,
where \emph{g} is a query representing the {\em goal}.
The query $g$ is evaluated against the knowledge base $\langle T,A_i$, where $A_i$ represents one of the states of the DKB.

Our aim is to build a {\em planning graph}~$P$ representing all possible plans.
The planning graph $P$ is a set of tuples~$\langle A, \acname{a}, \vartheta \rangle$ composed of an \abox~$A$, an action~\acname{a} in~$\Gamma$, and a substitution~$\vartheta$ used to obtain the action instance~$\acname{a}\vartheta$.
A tuple~$\langle A, \acname{a}, \vartheta \rangle\in P$ represents the transition~\acdyn{\acname{a},\vartheta}{$A$}{$A'$} in the system.
Given a planning graph $P$, a plan $p$ is a path starting from the initial state~$A_0$ and terminating in a {\em goal state}, which is any state~$A'$ such that~\mbox{$\answ(q,T,A')\neq\emptyset$},
by using transitions (tuples~$\langle A, \acname{a}, \vartheta \rangle$) in $P$.
So, a plan $p$ is a sequence (\acname{a_1}$\vartheta_1$, \ldots , \acname{a_n}$\vartheta_n$) of actions instantiations.
As discussed in  Sec.~\ref{sec:Description Logic based dynamic systems}, actions could lead to inconsistent states (i.e.~\abox-es w.r.t.~the \tbox~$T$), 
thus we need to take into account this aspect in the construction of the planning algorithms.

As said before, by default we want to find all possible plans that are an answer to the planning problem.
Such choice is justified by the contest in which we want to operate, namely business rich domains, where the decision about which plan to follow could be demanded to other components than the planner itself.
Another reason, more related to our implementation, is that our actions do not have any cost function attached to them, thus making it risky to choose one plan other another basing the decision on classic metrics like the number of actions.
Anyway, the planning algorithms shown below can be easily parametrized to find all or just one plan.

\subsection{Forward Planning}

{\em Forward Planning} starts from an initial state and expands it by applying all executable actions in that state.
New states obtained in this way are added to the states that must be explored, except for those that are inconsistent or are goal states.
Actions are executable in a state if their guard has at least one answer in that state. For a given state and action, we have as many outgoing edges
as the number of answers. The {\em Forward Planning Algorithm} (FP) is presented in a pseudo-code fashion in Algorithm~\ref{alg:Forward Planning Algorithm}.

The algorithm FP takes as input a dynamic knowledge base $\langle T,A_0,\Gamma\rangle$ and a goal~$g$, representing the property of state to be reached,
and returns a planning graph {\em P} as output. FA operates on two sets of states: {\em R} (the {\em remaining} states), containing states that have
to be expanded, and {\em V} (the {\em visited} states), containing states that have already been expanded.
{\em R} is initialized to $\{A_0\}$, which is the initial state. Recall that states are represented by mentioning explicitly only the \abox,
since the {\em TBox} is constant.

The main loop of the algorithm takes one state from {\em R}, adds it to {\em V}, checks if {\em A} is consistent, and if so, proceeds to
expand that state. If~$A$ is inconsistent it is simply eliminated, but kept in the set of visited states to avoid further consistency tests over it.
We assume to have a function {\tt Consistent}, with boolean return value, that allows us to test the consistency of an \abox~w.r.t.~a~\tbox.
On elimination of~$A$, also the edges in~$P$ that lead to~$A$ must be eliminated, which is done by using the auxiliary function {\tt EdgesTo},
defined as follows:\\[1mm]
\makebox[42mm][r]{{\tt EdgesTo}($P$,\,$\Gamma$,\,$A$)}$~~=~\{ \langle A', \acname{a}, \vartheta \rangle \mid \langle A', \acname{a}, \vartheta \rangle \in P~\wedge~($\acshort{a}{q}{e}$) \in \Gamma~\wedge~A=A'\cup \{e\vartheta\} \}$\\[1mm]
This function removes from~$P$ the edges that can cause the generation of the inconsistent state~$A$.
On the contrary, if $A$ is a consistent state and also a goal state ($\answ(q,T,A')\neq\emptyset$) it is not expanded further.
Otherwise, if~$A$ is not a goal state, it is expanded by finding {\em all} of the outgoing edges as well as the corresponding
arrival states, which is done using the auxiliary function {\tt Next}, defined as follows:\\[1mm]
\makebox[42mm][r]{{\tt Next}(T,A,$\Gamma$)}$~~=~\{\langle \acname{a}, \vartheta, A'\rangle \mid ($\acshort{a}{$q$}{$e$}$) \in \Gamma \wedge \vartheta \in $ \answ($q,T,A$)$~\wedge~$\acdyn{\acname{a},\vartheta}{$A$}{$A'$}$\}$\\[1mm]
This function considers every action~\acshort{a}{$q$}{$e$} and every action instance obtained by considering an answer~$\vartheta$
in the set~\answ($q,T,A$), where~$q$ is the action guard.
For every such action~\acname{a} and answer~$\vartheta$, it expands~$A$ into the new state~$A'$ by the action~\acdyn{\acname{a},\vartheta}{$A$}{$A'$}.
The new states obtained through the function {\tt Next} are added to $R$, except for those that have already been visited (i.e.~the states in $V$).
The consistency of these new states will be tested when they will be selected for expansion from the set~$R$. 
Finally, the {\em Planning Graph}~$P$ is updated by adding the new edges found through the function {\tt Next}.

\begin{algorithm}[!ht]
\small
\DontPrintSemicolon
\SetKwFunction{Ans}{Ans}
\SetKwFunction{Consistent}{Consistent}
\SetKwFunction{EdgesTo}{EdgesTo}
%\SetKwFunction{Reasoning}{Reasoning}
\SetKwFunction{Next}{Next}
\SetKwInOut{Input}{input}
\SetKwInOut{Output}{output}
\Input{A dynamic knowledge base~$\langle T,A_0,\Gamma\rangle$ and a goal~$g$, with~$A_0$ consistent w.r.t.~$T$
%Actions = {($q_1,e_1$), ..., ($q_m,e_m$)}\\
%%SWRLRules = {($B_{m+1},H_{m+1}$), ..., ($B_n,H_n$)}\\
%G = {$g_1 , ... , g_n$}\\
%$\langle T,A_0 \rangle$
}
\Output{A planning graph~$P$}
\BlankLine
\Begin{
\makebox[102mm][l]{$P \asgn \emptyset$}(planning graph)\;
\makebox[102mm][l]{$R \asgn \{ A_0 \}$}(remaining states)\;
\makebox[102mm][l]{$V \asgn \emptyset$}(visited states)\;
\While{$R \neq \emptyset$}{
	$A \in R$\;
	$R \asgn R \setminus \{ A \}$\;
	$V \asgn V \cup \{ A \}$\;
	
	%\Reasoning{TBox,ABox}\;
	
	\uIf{$\neg\,$\Consistent{A,T}}{
		\makebox[90mm][l]{$P \asgn P~\setminus$~\EdgesTo($P$,\,$\Gamma$,\,$A$)}(remove edges reaching $A$)
	}
	\uElseIf{$\answ(g,T,A) \neq \emptyset$}{
		\textbf{skip}%/break
	}
	\Else{
%		\makebox[90mm][l]{$N \asgn \Next{T,A,$\Gamma$}$}(find outgoing edges) \;
		\makebox[90mm][l]{$R \asgn R~\cup~(\,\{ A' \mid \langle \acname{a}, \vartheta, A' \rangle \in \Next{T,A,$\Gamma$} \}\,/~V\,)$}(add new states to $R$)\;
		\makebox[90mm][l]{$P \asgn P~\cup~\{ \langle A, \acname{a}, \vartheta \rangle \mid \langle \acname{a}, \vartheta, A'\rangle \in \Next{T,A,$\Gamma$}\}$}(update plan)
	}
}
}
\caption{Forward Planning Algorithm\label{alg:Forward Planning Algorithm}}
\end{algorithm}

Note that the FP algorithms finds {\em all} possible plans, because the retrieval of a goal state does not interrupt the execution
of the loop, which terminates only when all states have been visited. The variant of the algorithm where only {\em one} plan is found
can be easily obtained by replacing the \textbf{skip} command, after the test on being a goal state, with a \textbf{break} command,
which interrupts the execution of the loop thus skipping the visit of all of the remaining states.

In both cases the FP algorithm always terminates because the set of all possible states (\abox-es) is finite, 
since (i)~the set of all possible individuals is finite and consists in the~\adom($A_0$)
plus the individuals occurring in the set of actions~$\Gamma$, (ii)~the set of all possible concepts and roles is finite as well,
and (iii)~actions cannot create new individuals, nor new predicate symbols.

As usual in graph search algorithms, different search strategies (such as breadth-first or depth-first) as well as heuristics can be added
by imposing a data structure on the set $R$ and a specific extraction strategy. Thus the algorithm FP can easily be adapted for an {\em informed} search.

The forward algorithm has drawbacks due to the fact that, being uninformed, it explores all feasible plans.
In particular, it produces also {\em redundant plans}, that are plans containing proper sub-plans.
We define a plan (\acname{a_1}$\vartheta_1$)\ldots(\acname{a_n}$\vartheta_n$) 
to be \emph{redundant} if there exists a proper sub-sequence of
(\acname{a_1}$\vartheta_1$)\ldots(\acname{a_n}$\vartheta_n$) which is itself a plan.
An example of redundant plan is discussed below.
\begin{example}\label{ex:planning,forward algorithm}
Let us consider the Dynamic Knowledge Base~$\langle T_0, A_0, \Gamma_0 \rangle$ where:\\[1mm]
\idt$T_0~=~T'$\\[1mm]
\idt$A_0~=~A \cup \{\cls{Manager}(\cons{e001}), \cls{TechnicalDoc}(\cons{d001}), \rol{canManage}(\cons{e002},\cons{d001}) \}$ \\[1mm]
\idt$\Gamma_0~=~\Gamma\cup\{$\action{sayHello}{x,y}{$\cls{Manager}(x)\wedge \cls{Technician}(y)$}{$\cls{Greeted}(y)$}$\}$\\[1mm]
The knowledge base $\langle T', A \rangle$ is taken from Example~\ref{ex:kb-simple},
and $\Gamma$ contains the action $\acname{appoint}$ defined in Example \ref{ex:action}). We are given the goal:~~
$g\leftarrow\rol{assignedTo}(\cons{d001}, \cons{e002})$\\[1mm]
The FA would produce the following steps:\\
(1) select $A_0$ as the first state to expand (it is consistent, but it is not a goal state since  $\answ(g,T_0,A_0) = \emptyset$),\\
(2) compute $\langle \acname{appoint}, \{ x \mapsto \cons{e001}, y \mapsto \cons{e002}, z \mapsto \cons{d001} \}, A_1 \rangle$,
where $A_1 = A_0 \cup \{ \rol{assignedTo}(\cons{d001}, \cons{e002})\}$, and
$\langle \acname{sayHello}, \{ x \mapsto \cons{e001}, y \mapsto \cons{e002} \}, A_2 \rangle$, 
where $A_2 = A_0 \cup \{ \cls{Greeted}(\cons{e002})\}$, using the function {\tt Next},\\
(3) update the set $R$ and the planning graph $P$ accordingly,\\
(4) select from $R$ the state $A_1$, which is consistent and is also a goal state, thus is skipped,\\
(5) expand the state $A_2$, which is consistent, but $\answ(g,T_0,A_2) = \emptyset$, so it is not a goal state),\\
(6) compute $\langle \acname{appoint}, \{ x \mapsto \cons{e001}, y \mapsto \cons{e002}, z \mapsto \cons{d001} \}, A_3 \} \rangle$, 
where $A_3 = A_2 \cup \{\rol{assignedTo}(\cons{d001}, \cons{e002})\}$, using the function {\tt Next}\\
(7) update the set $R$ and the planning graph $P$ accordingly,\\
(8) select from $R$ the state $A_3$, which is consistent and is also a goal state, thus it is skipped,\\
(9) terminate because $R$ is empty.\\
\noindent We show a graphical representation of the planning graph in Fig. \ref{fig:Forward Planning Algorithm - Planning Graph example} (for simplicity, we do not mention
the substitutions). The set of plans represented by $P$ is very simple: $\{(\acname{sayHello},\acname{appoint}),(\acname{appoint})\}$.
In particular, the plan $(\acname{sayHello},\acname{appoint})$ is redundant, since it contains the proper sub-plan $(\acname{appoint})$.
\begin{figure}[t]
\vspace{-5mm}
\centering
\begin{tikzpicture}[->,auto,>=latex',node distance=3cm,on grid,auto, decoration=snake, scale=\ImgScale, every node/.style={transform shape}] 
\tikzstyle{state}=[circle,draw, font=\small] 
\tikzstyle{description}=[node distance=1cm, align=center, font=\small, text width=6em]
\tikzstyle{action description}=[node distance=2cm, align=center, font=\small, text width=6em]

\node[state, initial] (S1) {S1};
\node[state] (S2) [right of= S1] {S2};
\node[state, accepting] [right of= S2] (S3) {S3}; 
\node[state, accepting] [below of= S2, node distance=1.5cm] (S4) {S4};

\node[description] (S1d) [above of=S1] {$A_0$};
\node[description] (S2d) [above of=S2] {$A_0 \cup \{ \cls{Greeted}(\cons{e002}) \}$};
\node[description] (S3d) [right of=S3, node distance=2cm] {$A_0 \cup $ $\{ \cls{Greeted}(\cons{e002}), $ $\rol{assignedTo}(\cons{d001}, \cons{e002}) \}$};
\node[description] (S4d) [right of=S4, node distance=2cm] {$A_0 \cup \{ \rol{assignedTo}(\cons{d001}, \cons{e002}) \}$};

%\node[action description] (S1a) [below right of=S1] {$\acname{sayHello}$};
%\node[action description] (S2a) [below right of=S2] {$\acname{appoint}(x,y,z): x \rightarrow \cons{e001}$; $y \rightarrow \cons{e002}$; $z \rightarrow \cons{d001}$};
%\node[action description] (S3a) [below right of=S3] {$\acname{review}(x,y): x \rightarrow \cons{d001}$; $y \rightarrow \cons{e002}$};

\path 
(S1) edge[action description, below, decorate] node {$\acname{sayHello}$} (S2) 
(S2) edge[action description, below, decorate] node {$\acname{appoint}$} (S3) 
(S1) edge[action description, bend right, left, decorate] node {$\acname{appoint}$} (S4)

(S1d) edge[dashed, -] node {} (S1)
(S2d) edge[dashed, -] node {} (S2)
(S3d) edge[dashed, -] node {} (S3)
(S4d) edge[dashed, -] node {} (S4)
; 
\end{tikzpicture} 
\caption{Forward Algorithm example, Planning Graph} \label{fig:Forward Planning Algorithm - Planning Graph example}
\vspace{-5mm}
\end{figure}
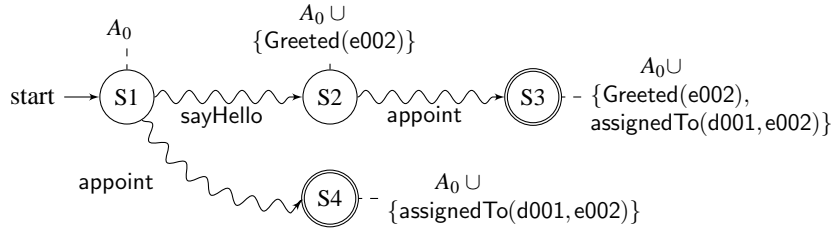
\end{example}

On large examples, the impact of redundant plans is a much larger planning graph.
The backward planning technique we are going to present in the rest of this section has the purpose of reducing the number of redundant plans found
and, thus, the size of the planning graph. The application of the Forward Planning Algorithm to the Case Study (Sec.~\ref{sec:Case Study})
is discussed in Sec.~\ref{sec:Experiments}.

\subsection{Backward State-space Reduction and Planning}

%\vtodo{The paper is generally well ­written, even though some sections are a
%bit difficult to understand, for instance the first paragraphs of
%section 4.2. \\
%I must admit that I have not really fully understood why the two ­stage
%scheme proposed here is more efficient than the one ­stage approach. I
%can see it in the description of the case study, but I am not sure I
%really understand the deeper reason why this works. The authors should
%describe this more carefully. \\
%The authors claim that the drawback of the forward algorithm is that
%it needs to explore all feasible paths. But isn't that just true if
%the algorithm proceeds to search for more plans after one has been
%found? How practically relevant is it really to find all plans and not
%just one? \\
%It should be worked out better what the abstraction is that is being
%computed as part of the ABP. What is the concretization function
%related to this abstraction? Does this abstraction have any particular
%property, for instance that of forming a Galois connection?}

{\em Backward Planning}~\cite{Ghallab2004a} is based on the idea of starting from a goal state and, analysing actions that can lead to such state, build the set of predecessor states until either the initial state is found or the state-space is explored entirely, thus being {\em goal-driven}.
In lack of specific information on the state-space structure, \emph{backward algorithms} can have advantages over forward algorithms since they explore a smaller portion of the state-space by considering only actions that can lead to the goal satisfaction.
On the contrary, in the forward approach and under uninformed search, all the executable actions are taken into account and more states are generated.
% In practice, however, it is hard to predict which approach will perform best. 
%It is common to combine the two approaches into various forms of bi-directional search~\cite{Russell2010}, to take advantage of the two approaches.
Another very common and widely used technique to make the planning problem easier to solve is to create an \emph{abstraction} of it~\cite{Russell2010}: the original planning domain is shrank by omitting details, thus reducing the size of it (by removing superfluous states) and making it more manageable, but without loosing the capability of finding plans.

This is also what we do in the proposed {\em Backward State-space Reduction} technique, where we combine abstraction and backward state-search.
The idea is to structure the planning in two phases (identified jointly as ABP+FPI): 
first, an {\em Abstract Planning Graph}~$\mathcal{P}$ is created by using the {\em Abstract Backward Planning Algorithm} (ABP),
then the abstract planning graph~$\mathcal{P}$ is instantiated into a corresponding planning graph $P$ by using the {\em Forward Plan Instantiation Algorithm}~(FPI), which is essentially a variant of FP.
%
%we therefore have less states to explore (since we take into consideration less actions),
%and the plans we find don't contain redundant actions (as it could happen in the FA).
%
The abstraction we apply in ABP is that the algorithm manipulates states represented by {\em queries} rather than \abox-es.
A query~$q$ (which we call an {\em abstract state}) represents a set~\concrete{$q$,$T$} of \abox-es such that~$A\in~$\concrete{$q$,$T$} iff~
(i)~$A$ is consistent w.r.t.~$T$ and
(ii)~$\answ(q,T,A)\neq\emptyset$.
In this sense, the algorithm is {\em symbolic} and manipulates sets of (possible) states rather than single states.
%Such abstraction, though, is not the classical abstraction used in planning as it is not a surjective function:
%this is due to the fact that an $\abox$ could be represented by one or more queries (thus abstract states), and not only one.
%Nonetheless, using such an abstraction helps us in building an abstract planning domain which results more compact and easy to reason with than the original one (see later for examples).
%
Having the abstract planning domain, the ABP produces (through a backward search) an {\em abstract planning graph}~$\mathcal{P}$, which is a set of tuples~$\langle \sigma, \acname{a} \rangle$ composed of an abstract state~$\sigma$ and an action~\acshort{a}{$q$}{$e$} in~$\Gamma$. 
The tuple~$\langle \sigma, \acname{a} \rangle \in P$ represents {\em the set of all transitions}~\acdyn{\acname{a},\vartheta}{$A$}{$A'$}, such that
(i)~$A\in~$\concrete{$\sigma$,$T$}, and
(ii)~$\vartheta\in\answ(q\wedge\sigma,T,A)$.
In other words, we interpret~$\langle \sigma, \acname{a} \rangle$ as a constraint over the guard~$q$ of action~\acname{a}, which is refined using the abstract state (i.e.~query)~$\sigma$.

The abstract planning graph~$\mathcal{P}$ is then used within the~FPI algorithm, which is essentially a variant of FP, as a pattern to direct the search of the concrete plan $P$.
In particular, we force the choice of actions to be applied and the actions have also stronger guard, as we shall discuss in the following.
The advantage of enforcing constraints over actions, w.r.t.~the FP algorithm, is to significantly reduce the branching of the planning by exploiting information propagated from the goal condition in a backward manner.
This can be seen as a kind of informed search where the preliminary abstract backward phase is useful to discover properties of the state-space that can be used to direct the second phase.

\begin{algorithm}[!ht]
\small
\DontPrintSemicolon
\SetKwFunction{Ans}{Ans}
\SetKwFunction{Resolution}{Resolution}
\SetKwFunction{Resolve}{FullyResolve}
\SetKwFunction{PrevA}{PrevA}
\SetKwInOut{Input}{input}
\SetKwInOut{Output}{output}
\Input{A dynamic knowledge base~$\langle T,A_0,\Gamma\rangle$ and a goal~$g$\\
% Actions = {($q_1,e_1$), ..., ($q_m,e_m$)}\\
%SWRLRules = {($q_{m+1},e_{m+1}$), ..., ($q_n,e_n$)}%\\
%G = {$g_1 , ... , g_n$}\\
%$\langle T,A_0 \rangle$
}
\Output{An abstract planning graph $\mathcal{P}$}
\BlankLine
\Begin{
\makebox[5mm][l]{$\mathcal{P}$}\makebox[85mm][l]{$\asgn~\emptyset$}(abstract planning graph)\;
\makebox[5mm][l]{$R$}\makebox[85mm][l]{$\asgn~\{g\}$}(remaining abstract states)\;
\makebox[5mm][l]{$V$}\makebox[85mm][l]{$\asgn~\emptyset$}(visited abstract states)\;
%\Reasoning{TBox,$ABox_0$}\;

\While{$R \neq \emptyset$}{
	$\sigma \in$ R\;
	\makebox[4mm][l]{$R$}$\asgn~R\,\setminus \{\sigma\}$\;
	\makebox[4mm][l]{$V$}$\asgn~V \cup \{\sigma\}$\;	
	\makebox[85mm][l]{}(test if $\sigma$ is satisfied in the initial state)\\[-4.5mm]
	\lIf{\rm\answ($\sigma,T,A_0$) $\neq \emptyset$}{\;
 	    \idt \textbf{skip}%/break
	}
	\makebox[85mm][l]{}(apply simple join axioms)\\[-4.5mm]
	\For{$\sigma'$ {\bf in} \Resolve{$\sigma,\sj(T)$}}{\vspace{1mm}
%	  \For{{\rm($q$,$e$) {\bf in} $\Gamma$}}{
%		\For{ $(\sigma',\vartheta)$ {\rm\bf in} \Resolution{\rm ($q$,$e$),$\sigma$}}{
%			\makebox[5mm][l]{$\varphi'$}$\asgn~(\sigma \setminus e\vartheta)\wedge q\vartheta$\;
			\makebox[5mm][l]{$R$}\makebox[74mm][l]{$\asgn~R~\cup\,(\,\{\sigma''\mid \langle\sigma'',\acname{a}\,\rangle\!\in~$\PrevA{$\sigma',\Gamma$}$\}\setminus\,V\,)$}(update remaining states)\;\vspace{1mm}
			\makebox[5mm][l]{$\mathcal{P}$}\makebox[74mm][l]{$\asgn~\mathcal{P}~\cup~\,$\PrevA{$\sigma',\Gamma$}}(update plan)
			%\{\,\langle \sigma'', \acname{a}, \vartheta \rangle\ \mid\ (\sigma'',\acname{a},\vartheta)\!\in\,$\PrevA{$\sigma',\Gamma$}$\,\}$}(update plan)
%		}
%	  }
	}
}
}
\caption{Abstract Backward Planning Algorithm\label{alg:Backward Planning Algorithm - Abstract Plan creation}}
\end{algorithm}

%k. io direi perche' nel forward lavoriamo con ABox complete e, controllandone la consistenza applichiamo anche gli assiomi
%[8:31:29 PM] Michele Stawowy: nel backward invece usiamo formule, e quindi dobbiamo "prevedere" se nella relativa abox
%[8:31:39 PM] Michele Stawowy: verranno applicati gli assiomi o meno
%[8:31:55 PM] Michele Stawowy: senza fare questa previsione, rischiamo di perdere dei possibili piani

We now describe the two algorithms in detail. The Abstract Backward Planning Algorithm is given in pseudo-code
in Algorithm~\ref{alg:Backward Planning Algorithm - Abstract Plan creation}.
It keeps track of remaining abstract states to be explored
in the set~$R$ and of visited abstract states in the set~$V$. For every abstract state~$\sigma$ in~$R$, the algorithm tests whether it includes
the initial \abox~$A_0$, if so the abstract state is not further expanded. Otherwise, the abstract state~$\sigma$ is {\em resolved} to a new set of abstract
states by applying as much as possible the simple join axioms contained in the \tbox~$T$ (they are denoted by~$\sj(T)$). This step is performed
using the auxiliary function:\\[1mm]
\makebox[42mm][r]{{\tt FullyResolve}($\sigma,\sj(T)$)~~}$=$\\[1mm]
\makebox[22mm][l]{}$~\{\,\sigma'\,\mid\,\sigma'$ obtained by applying~{\tt Resolve} to~$\sigma$ as much as possible~using~$\sj(T)$ axioms$\,\}$\\[1mm] %\{\langle \acname{a}, \vartheta, A'\rangle \mid ($\acshort{a}{$q$}{$e$}$) \in \Gamma \wedge \vartheta \in $ \answ($q,T,A$)$~\wedge~$\acdyn{\acname{a},\vartheta}{$A$}{$A'$}$\}\}$\\[1mm]
which relies on the {\tt Resolve} function, that applies a form of SLD-resolution in the spirit of Logic Programming~\cite{Lloyd:1993:FLP:529834}.
In particular, a new abstract state~$\sigma'$ is obtained from the abstract state~$\sigma$ using an 
axiom~$\cls{N}_1(x)\wedge\cls{N}_2(y) \rightarrow \rol{R}(x,y)$ in~$\sj(T)$ and replacing 
a conjunct of the form~$(a_1\wedge\ldots\wedge\rol{R}(x,y)\wedge\ldots\wedge a_n)\vartheta$ in~$\sigma$, for some substitution~$\vartheta$,
with the corresponding conjunct~$(a_1\wedge\ldots\wedge\cls{N}_1(x)\wedge\cls{N}_2(y)\wedge\ldots\wedge a_n)\vartheta$.
In principle, we may have many axioms with the same conclusion and this is the reason why the function {\tt FullyResolve}
returns a {\em set} of possible resolved abstract states.
The importance of this step is due to the fact that atoms in the conclusions of simple axioms have disjoint predicates 
from atoms in the effect of the actions: by resolving~$\sigma$ w.r.t.~simple join axioms we enable
the (backward) application of more actions than those applicable directly in~$\sigma$.
On the contrary, in forward planning we have a concrete \abox, the set of applicable actions
depends on that \abox, and the simple join axioms do not compare explicitly in the algorithm as they
are used, implicitly and together with the other axioms in the \tbox, to test the satisfiability of the action guard.

%, unless we are considering abstract states that already include the initial \abox~$A_0$.
% THIS IS IMPOSSIBLE BECAUSE IN THE PREVIOUS TEST ON ANS THIS CHECK IS INCLUDED
For every resolved abstract state~$\sigma'$ the algorithm computes the set of previous abstract states
by considering all possible actions in~$\Gamma$ that have an effect that is (unifiable with) an atom~$a$
in~$\sigma'$ and replacing~$a$ with the corresponding (unified) action guard. This is performed by the
auxiliary function:\\[1mm]
         \makebox[42mm][r]{{\tt        PrevA}($\sigma,\Gamma$)~~}$=~\{\,\langle \sigma', \acname{a}\rangle\,\mid~($\acshort{a}{$q$}{$e$}$) \in \Gamma~\wedge~\sigma'\in\,{\tt ActPrevA}(\sigma,($\acshort{a}{$q$}{$e$}$))\,\}$\\[1mm]
         \makebox[42mm][r]{{\tt     ActPrevA}($\sigma,($\acshort{a}{$q$}{$e$}$)$)~~}$=~\{\,\sigma'\,\mid~\sigma'$ obtained~by applying~{\tt Resolve} to~$\sigma$ using~\acshort{a}{$q$}{$e$}$\,\}$\\[1mm]
         %={\tt Repl}((a_1\wedge\ldots\wedge e\wedge\ldots a_n)\vartheta,(a_1\wedge\ldots\wedge q\wedge\ldots a_n)\vartheta,\sigma) \,\}$\\[1mm]
%\mathit{conj}\in\sigma \wedge \mathit{conj}=(a_1\wedge\ldots\wedge a_n) \wedge (a_i=e)\vartheta \wedge \sigma'\in\text{\tt Replace}(conj,(a_1\wedge\ldots\wedge a_n)\vartheta,\sigma)\}$\\[1mm] %%,(a_1\wedge\ldots\wedge a_n)\vartheta,\sigma
%\vartheta \in $ \answ($q,T,A$)$~\wedge~$\acdyn{\acname{a},\vartheta}{$A$}{$A'$}$\}\}$\\[1mm]
which, again, relies on the {\tt Resolve} function that computes the new abstract state~$\sigma'$
by using the action~\acshort{a}{$q$}{$e$} and replacing a conjunct of the form~$(a_1\wedge\ldots\wedge e\wedge\ldots\wedge a_n)\vartheta$
in~$\sigma$, for some substitution~$\vartheta$, with the corresponding conjunct~$(a_1\wedge\ldots\wedge q\wedge\ldots\wedge a_n)\vartheta$.

Termination of the ABP algorithm follows from the fact that we can have finitely many abstract states (i.e.~queries),
for similar reasons to those discussed for termination of the FP algorithm.
\begin{example}\label{ex:planning,abstract backward algorithm}
Let us consider the DKB and the goal defined in Example \ref{ex:planning,forward algorithm}. \\
The ABP would produce the following steps:\\
(1) the goal state~$g$ is not satisfied in the initial KB ($\answ(goal,T_0,A_0) = \emptyset$) and it is kept for expansion,\\
(2) the set~$\sj(T)$ of simple join axioms is empty, so the function {\tt FullyResolve} returns the state~$g$ itself,\\
(3) the function {\tt PrevA} computes the pair $\langle S_2, \acname{appoint} \rangle$, where $S_2=\cls{Manager}(x) \wedge \cls{Technician}(\cons{e002})$,\\
(4) the set $R$ and the abstract planning graph $\mathcal{P}$ are updated accordingly,\\
(5) $S_2$ is selected from $R$ and, since it is satisfied in the initial KB ($\answ(S_2,T_0,A_0)\neq \emptyset$), it is not expanded,\\
(6) the set $R$ is empty and the algorithm terminates.
We show the resulting abstract graph in Fig.~\ref{fig:Backward Planning Algorithm example}.
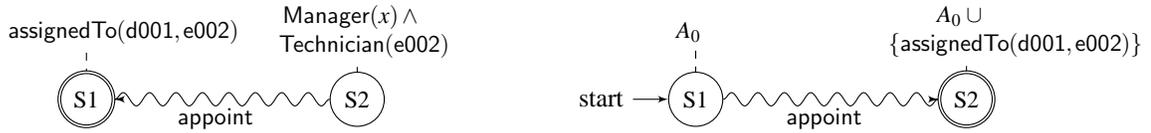
\begin{figure}[t]
\vspace{-4mm}
\centering
\begin{tikzpicture}[->,auto,>=latex',node distance=4cm,on grid,auto, decoration=snake, scale=\ImgScale, every node/.style={transform shape}] 
\tikzstyle{state}=[circle,draw, font=\small] 
\tikzstyle{description}=[node distance=1cm, align=center, font=\small, text width=6em]

\node[state, accepting] (S1) {S1};
\node[state] (S2) [right of= S1] {S2};

\node[description] (S1d) [above of=S1] {$\rol{assignedTo}(\cons{d001}, \cons{e002})$};
\node[description] (S2d) [above of=S2] {$\cls{Manager}(x) \wedge \cls{Technician}(\cons{e002})$};

\path 
(S2) edge[description, below, decorate] node {$\acname{appoint}$} (S1) 

(S1d) edge[dashed, -] node {} (S1)
(S2d) edge[dashed, -] node {} (S2)
;

\node[state, initial] (S1n) [right of= S2, node distance=5cm] {S1};
\node[state, accepting] (S2n) [right of= S1n] {S2};

\node[description] (S1nd) [above of=S1n] {$A_0$};
\node[description] (S2nd) [above of=S2n] {$A_0 \cup \{ \rol{assignedTo}(\cons{d001}, \cons{e002}) \}$};

\path 
(S1n) edge[description, below, decorate] node {$\acname{appoint}$} (S2n)

(S1nd) edge[dashed, -] node {} (S1n)
(S2nd) edge[dashed, -] node {} (S2n)
;

\end{tikzpicture} 
\caption{ABP+FPI example: Abstract Planning Graph (left) and Instantiated Planning Graph (right)} \label{fig:Backward Planning Algorithm example}
\vspace{-4mm}
\end{figure}
\end{example}

\begin{algorithm}[!ht]
\small
\DontPrintSemicolon
\SetKwFunction{Ans}{Ans}
\SetKwFunction{Reasoning}{Reasoning}
\SetKwFunction{Consistent}{Consistent}
\SetKwFunction{NextA}{NextA}
\SetKwInOut{Input}{input}
\SetKwInOut{Output}{output}
\Input{A dynamic knowledge base~$\langle T,A_0,\Gamma\rangle$, an abstract planning graph~$\mathcal{P}$, and a goal~$g$\\
%Actions = {($q_1,e_1$), ..., ($q_n,e_n$)}\\
%SWRLRules = {($q_{m+1},e_{m+1}$), ..., ($q_n,e_n$)}\\
%$P_\varphi$\\
%$\langle T,A_0 \rangle$
}
\Output{A planning graph~$P$}
\BlankLine
\Begin{
\makebox[102mm][l]{$P \asgn \emptyset$}(planning graph)\;
\makebox[102mm][l]{$R \asgn \{ A_0 \}$}(remaining states)\;
\makebox[102mm][l]{$V \asgn \emptyset$}(visited states)\;
%$\{ \langle A_0, (q,e), \vartheta \rangle \mid \langle \sigma, (q,e), \vartheta \rangle \in \mathcal{P}~\wedge~$\answ($\sigma,T,A_0$) $\neq \emptyset \}$\;
%$\{\, (A_0\cup\{e\vartheta\}) \mid \langle \sigma, (q,e), \vartheta \rangle \in \mathcal{P}~\wedge~$\answ($\sigma,T,A_0$) $\neq \emptyset \}$\;
%$\{A_0\}$\;
\While{$R \neq \emptyset$}{
	$A \in R$\;
	$R \asgn R \setminus \{ A \}$\;
	$V \asgn V \cup \{ A \}$\;
	
	%\Reasoning{TBox,ABox}\;
	\uIf{$\neg\,$\Consistent{A,T}}{
		\makebox[90mm][l]{$P \asgn P~\setminus$~\EdgesTo($P$,\,$A$)}(remove edges reaching $A$)
	}
	\uElseIf{$\answ(g,T,A) \neq \emptyset$}{
		\textbf{skip}%/break
	}
	\Else{
%		\makebox[90mm][l]{$N \asgn \Next{T,A,$\Gamma$}$}(find outgoing edges) \;
		\makebox[90mm][l]{$R \asgn R~\cup~(\,\{ A' \mid \langle \acname{a}, \vartheta, A' \rangle \in \NextA{T,A,$\mathcal{P},\Gamma$} \}\,/~V\,)$}(add new states to $R$)\;
		\makebox[90mm][l]{$P \asgn P~\cup~\{\, \langle A, \acname{a}, \vartheta \rangle \mid \langle \acname{a}, \vartheta, A'\rangle \in \NextA{T,A,$\mathcal{P},\Gamma$}\}$}(update plan)
	}

%	\uIf{$\neg$\,\Consistent{T,A}}{
%		$P \asgn P~\setminus$~\EdgesTo($P$,\,$A$)
%	}
%	\Else{
%		\For{$\langle (q,e), \vartheta \rangle \in \{ \langle (q,e), \vartheta \rangle \mid \langle \varphi, (q,e), \vartheta \rangle \in P_\varphi \}$}{
%			\lIf{(q,e) $\in$ SWRLRules}{\;
%			\idt$N \asgn \{ A \}$}
%			\lElse{\;
%			\idt$N \asgn$ \Next{T,A,\{(q,e)\}}}
%			\idt$\varphi' \asgn \varphi \cup e\vartheta$\;
%			\idt$R \asgn R \cup (\{ \langle A', \varphi' \rangle \mid ABox' \in N \}) \setminus V$\;
%			\idt$P \asgn P \cup \{ \langle A, (q,e), \vartheta_P \rangle \mid \langle (q,e), \vartheta_P, A'\rangle \in N \}$\;
%		}
%	}

}
}
\caption{Forward Plan Instantiation Algorithm \label{alg:Backward Planning Algorithm - Planning Graph concretization}}
\end{algorithm}

The Forward Plan Instantiation Algorithm is given in pseudo-code in Algorithm~\ref{alg:Backward Planning Algorithm - Planning Graph concretization}.
This algorithm is very similar to the FP algorithm: it takes as input a dynamic knowledge base, a goal, and also an abstract planning graph. 
The algorithm differs from FP in the fact that actions form~$\Gamma$ are executed under the constraints present in~$\mathcal{P}$.
The plan construction starts from the initial state~$A_0$ and each new state~$A$ in~$R$ is expanded by considering 
pairs~$\langle \sigma, \acname{a} \rangle \in \mathcal{P}$ such that the abstract state~$\sigma$ includes~$A$ and the action~\acname{a}
is executed under the extra precondition~$\sigma$. This is performed by the following auxiliary function:\\[1mm]
\makebox[42mm][r]{{\tt        NextA}($T,A,\mathcal{P},\Gamma$)~~}$=~\{ \langle \acname{a}, \vartheta, A' \rangle \mid \langle \sigma, \acname{a} \rangle \in \mathcal{P}\wedge~A\in~$\concrete{$\sigma$,$T$}$~\wedge$\\
\makebox[75mm][r]{}$~$(\acshort{a}{$q$}{$e$})$~\in\Gamma~\wedge~\vartheta\in\answ(q\wedge\sigma,T,A)~\wedge~\,$\acdyn{\acname{a},\vartheta}{$A$}{$A'$}$\,\}$
which consider effects~\acdyn{\acname{a},\vartheta}{$A$}{$A'$} over the current state~$A$ with an instantiation of the action~\acname{a}
that is computed in the set~$\answ(q\wedge\sigma,T,A)$ of answers restricted with the extra precondition~$\sigma$.

The termination of the FPI algorithm is granted by the same observations make for the FP algorithm.
Soundness is ensured by the fact that the algorithm find a (not-necessarily proper) subset of the plans
found by the FP algorithm. Concerning completeness, we are currently working on the notion of non-redundant
plan and on a relative completeness result, stating that the ABP+FPI algorithm is complete w.r.t.~the set of
non-redundant plans computed by the FP algorithm.

\noindent Let us now show the FPI algorithm at work.
\begin{example}\label{ex:planning, backward algorithm}
Consider the DKB and the goal defined in Example~\ref{ex:planning,forward algorithm} and 
the Abstract Planning Graph obtained in Example~\ref{ex:planning,abstract backward algorithm}.
The FPI produces the following steps:\\
(1) $A_0$ is taken as the first state to be expanded, since it is consistent but it is not a goal state,\\
(2) using the function {\tt NextA}, find the tuple $\langle \acname{appoint}, \{ x \mapsto \cons{e001}, y \mapsto \cons{e002}, z \mapsto \cons{d001} \}, A_1 \rangle$, where\linebreak $A_1 = A_0 \cup \{ \rol{assignedTo}(\cons{d001}, \cons{e002}) \}$\\
(3) the set $R$ and the planning graph $\mathcal{P}$ are updated accordingly,\\
(4) the state $A_1$ is selected from $R$, it is consistent w.r.t.~$T$ and it is a goal state ($\answ(q, T_0, A_1)$),
(5) the set $R$ is empty, so the algorithm terminates. 
We show the resulting graph in Fig. \ref{fig:Backward Planning Algorithm example}(for simplicity, we omit the substitution function $\vartheta$).
%\begin{figure}[t]
%\centering
%\input{imgs/img_bw_basic_example.tex}
%\caption{FPI example, Planning Graph} \label{fig:Backward Planning Algorithm - Planning Graph example}
%\vspace{-4mm}
%\end{figure}
\end{example}

\noindent As we can see from Example \ref{ex:planning, backward algorithm}, the ABP+FPI produces a smaller final graph if compared to
Figure~\ref{fig:Forward Planning Algorithm - Planning Graph example}.
%
%final graph in which there are no redundant plans.
%\vcomm{COMMENTS AND COMPARISON OF THE TWO possiamo dire che non produce piani ridondanti poiche, andando all'indietro, facciamo solo 
%sostituzioni di atomi. Se un atomo non è nel goal, allora non consideriamo le azioni che lo aggiungono.}
%The application of the Abstract Backward Planning Algorithm and Forward Plan Instantiation Algorithm 
%to the Case Study (Sec. \ref{sec:Case Study}) is presented in Sec. \ref{sec:Experiments}.

%---------------

%\vtodo{Vantaggi BA:\\
%	piu' performante e veloce??? Dobbiamo costruire e controllare la consistenza di meno KB, la funzione Next() usa una sola azione alla volta, non tutto il set}\\	

%\section{Planning in Dynamic Knowledge Bases - Old version}
%
%\input{tex/planning_ms.tex}

% -----------------

%\newpage
\section{Experiments}
\label{sec:Experiments}

%\vtodo{
%My main concern is the not very extensive
%experimental evaluation. The paper uses jst one example that is
%provided by the authors which is an issue since the authors may have
%provided a case study that is biased towards working well with the
%method that the authors themselves propose. It would be a lot more
%convincing if the authors had analyzed multiple case studies. Also,
%the planning community regularly holds meetings in which different
%planning tools are applied to various benchmark problems (the ICAPS
%competition) and this approach should be experimentally evaluated in
%this context. However, since we are considering this paper for a
%workshop, not a conference, this somewhat flimsy experimental
%evaluation may be considered acceptable.
%}
%
%\vtodo{Per risolvere l'osservazione fatta dal reviewer, cercare un esempio di problema usato nelle gare di planning ed eseguirlo. Non dovrebbe essere necessario spiegarlo più di tanto.}

% Solver, reasoner, librerie usate. Riferimento a Mastro, perche' non e' stato usato.
% Num stati e edge
% commenta il codice in generale, indipendenza dal reasoner
We have implemented the algorithms presented in Sec. \ref{sec:Planning in DLDSs} and in this section we report on the implementation
as well as on empirical results obtained by applying the algorithms to the case study of Sec.~\ref{sec:Case Study}.
The implementation, made with Python, provides the FP algorithm or the ABP+FPI algorithm, 
specifying for each of them the strategy preferred (depth- or breadth-first, only the first solution or all of them).

Since the DKB is based on DL-Lite fragment, we need a reasoner to check consistency and querying knowledge bases.
The reasoner Mastro~\footnote{\url{http://www.dis.uniroma1.it/~mastro/}} supports DL-Lite but it is still in closed beta
and works mainly as a stand alone system. Furthermore, Mastro does not support reasoning over Simple Join axioms.
Since DL-Lite is a subset of the \textit{Web Ontology Language} (OWL)~\cite{Calvanese2009}, we resort to
the the reasoner Pellet\footnote{\url{http://clarkparsia.com/pellet/}}, a popular and freely available OWL2 reasoner, that has all 
the features interested in. In particular, Pellet supports SJ axioms encoded as SWRL Rules~\cite{Sirin200751}.
%Since Pellet accepts ontologies written using OWL syntax, we translated DL-Lite axioms in OWL axioms.
Provided the reasoner satisfies our requirements, its choice is not crucial for our planning algorithm, 
since it is parametric w.r.t.~the chosen reasoner.

To test the two algorithms, we created various \abox-es differing only for the number of instances
and we varied the number of instances participating in the classes
$\cls{Manager}$, $\cls{Employee}$ and $\cls{TechnicalDoc}$.
This affects the size of the planning search space and it is useful to assess how the algorithms scale.

\begin{figure}[t]
\vspace{-3mm}
\centering
\begin{tikzpicture}[->,auto,>=latex',node distance=4cm,on grid,auto, decoration=snake, scale=\ImgScale, every node/.style={transform shape}] 
\tikzstyle{state}=[circle,draw] 
\tikzstyle{description}=[node distance=1.5cm, align=center, font=\small, text width=14em]
%\tikzstyle{action}=[node distance=1.5cm,draw, minimum height=2.5cm]
\tikzstyle{action description}=[node distance=0cm, align=center, font=\small, shift={(0.3 cm,0cm)}]

\node[state, initial] (S1) {S1};
\node[state] (S2) [right of= S1] {S2};
\node[state] [right of= S2] (S3) {S3}; 
\node[state, accepting] [right of= S3] (S4) {S4};

\node[description] (S1d) [above of=S1] {$A_0$};
\node[description] (S2d) [above of=S2] {$A_0 \cup \{ \cls{Technician}(\cons{e002}) \}$};
\node[description] (S3d) [above of=S3] {$A_0 \cup \{ \cls{Technician}(\cons{e002}) $, $ \rol{assignedTo}(\cons{d001},\cons{e002}) \}$};
\node[description] (S4d) [above of=S4] {$A_0 \cup \{ \cls{Technician}(\cons{e002}) $, $ \rol{assignedTo}(\cons{d001},\cons{e002}) $, $ \rol{hasStatus}(\cons{d001},\cons{reviewed}) \}$};

%\node[action] (S1a) [below right of=S1] {};
\node[action description,below right] at (S1.south east) {$\acname{setTechnician}(x,y): $ \\
$x \rightarrow \cons{e001}$; \\
$ y \rightarrow \cons{e002}$};
%\node[action] (S2a) [below right of=S2] {};
\node[action description,below right] at (S2.south east) {$\acname{appoint}(x,y,z):$\\
$x \rightarrow \cons{e001}$;\\
$y \rightarrow \cons{e002}$;\\
 $z \rightarrow \cons{d001}$};
%\node[action] (S3a) [below right of=S3] {};
\node[action description,below right] at (S3.south east) {$\acname{review}(x,y):$\\
$x \rightarrow \cons{d001}$;\\
$y \rightarrow \cons{e002}$};

\node[] (filling) [below of=S1, node distance=3cm] { };

\path 
(S1) edge[decorate] node {} (S2) 
(S2) edge[action description, below, decorate] node {} (S3) 
(S3) edge[action description, below, decorate] node {} (S4)

(S1d) edge[dashed, -] node {} (S1)
(S2d) edge[dashed, -] node {} (S2)
(S3d) edge[dashed, -] node {} (S3)
(S4d) edge[dashed, -] node {} (S4)
; 
\end{tikzpicture}
\vspace{-14mm}
\caption{Planning Graph} \label{fig:Forward Planning Algorithm - Planning Graph}
\vspace{-3mm}
\end{figure}

In Fig.~\ref{fig:Forward Planning Algorithm - Planning Graph} we present the Planning Graph obtained with FP 
(the instantiated planning graph obtained with ABP+FPI is identical), considering 1 manager, 1 employee and 1 technical document.
For this small problem instance the two algorithms (FP and ABP+FPI) take the same amount of time.
This can be explained by looking at the column $Inc$ in Table \ref{tab:Empirical Results}, counting the number of inconsistent states 
that each the algorithm finds. Even for such a simple example, FP finds 13 inconsistent states, while FPI, thanks to the constrains 
provided by the Abstract Planning Graph, finds only 3 inconsistent states.

\begin{figure}[!ht]
\centering
\begin{tikzpicture}[->,auto,>=latex',node distance=1.5cm,on grid,auto, decoration=snake, scale=\ImgScale, every node/.style={transform shape}] 
\tikzstyle{state}=[circle,draw] 
\tikzstyle{state initial}=[circle,fill=gray!20,draw] 
\tikzstyle{description}=[node distance=4cm, align=center, font=\small]
	
\node[state, accepting] (S1) {S1};
\node[state] (S2) [below of= S1] {S2};
\node[state] [below of= S2] (S3) {S3}; 
\node[state] [below left of= S3,node distance=2cm] (S4) {S4}; 
\node[state] [below right of= S3,node distance=2cm] (S5) {S5}; 
\node[state initial] [below of= S4] (S6) {S6}; 
\node[state initial] [below of= S5] (S7) {S7}; 

\node[description] (S1d) [right of=S1] {$\rol{hasStatus}(x,\cons{reviewed}) \wedge \cls{UrgDoc}(x)$};
\node[description] (S2d) [right of=S2] {$\rol{assignedTo}(x,y) \wedge \cls{UrgDoc}(x)$};
\node[description] (S3d) [right of=S3] {$\cls{Mng}(x) \wedge \rol{cm}(y,z)  \wedge \cls{UrgDoc}(z)$};
\node[description] (S4d) [left of=S4, text width=14em] {$\cls{Mng}(x) \wedge \cls{Tech}(y) \wedge \cls{TDoc}(z) \wedge \cls{UrgDoc}(z)$};
\node[description] (S5d) [right of=S5] {$\cls{Mng}(x) \wedge \cls{Adm}(y) \wedge \cls{ADoc}(z) \wedge \cls{UrgDoc}(z)$};
\node[description] (S6d) [left of=S6, text width=14em] {$\cls{Mng}(w) \wedge \cls{Mng}(x) \wedge \cls{Empl}(y) \wedge \cls{TDoc}(z) \wedge \cls{UrgDoc}(z)$};
\node[description] (S7d) [right of=S7, text width=14em] {$\cls{Mng}(w) \wedge  \cls{Adm}(z) \wedge \cls{Mng}(x) \wedge \cls{Doc}(y)  \wedge \cls{UrgDoc}(y)$};

\path 
(S2) edge[description, right, decorate] node {$\acname{review}[x,y]$} (S1) 
(S3) edge[description, right, decorate] node {$\acname{appoint}[x,y,z]: z \rightarrow x$} (S2) 
(S4) edge[description, left] node {SJ axiom} (S3) 
(S5) edge[description, right] node {SJ axiom} (S3) 
(S6) edge[description, left, decorate] node {$\acname{setTechnician}[x,y]: w \rightarrow x$} (S4) 
(S7) edge[description, right, decorate] node {$\acname{setAdmDoc}[x,y]: y \rightarrow z, w \rightarrow x, z \rightarrow y$} (S5) 

(S1d) edge[dashed, -] node {} (S1)
(S2d) edge[dashed, -] node {} (S2)
(S3d) edge[dashed, -] node {} (S3)
(S4d) edge[dashed, -] node {} (S4)
(S5d) edge[dashed, -] node {} (S5)
(S6d) edge[dashed, -] node {} (S6)
(S7d) edge[dashed, -] node {} (S7)
; 
\end{tikzpicture} 
\caption{Abstract Planning Graph} \label{fig:Backward Planning Algorithm - Abstract Graph}
\end{figure}
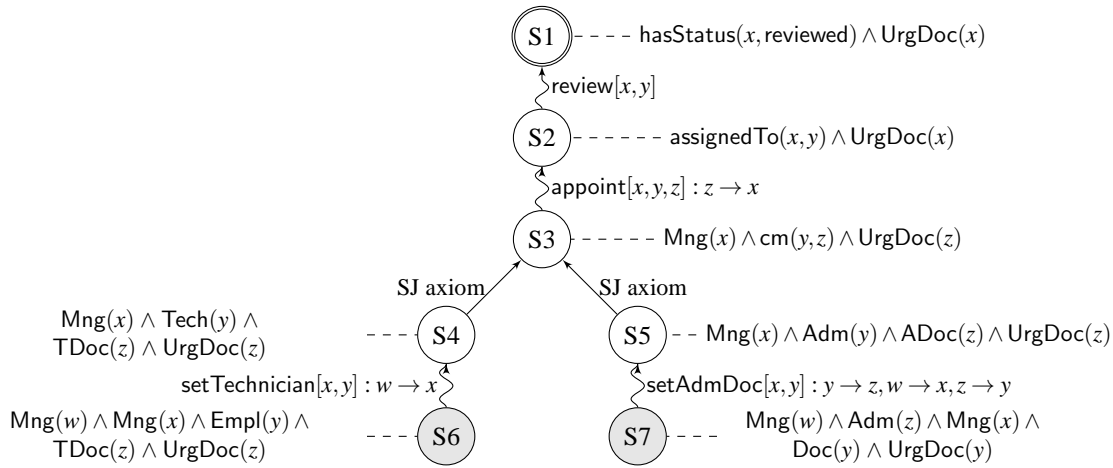

In Fig.~\ref{fig:Backward Planning Algorithm - Abstract Graph} we show the \textit{Abstract Planning Graph} obtained by the ABP 
(gray states are initial states). The Simple Join axioms applications are explicitly shown to make it easier to understand the algorithm behaviour.
The graph is {\em constant} for all the \abox-es we created, because no matter what is the number of instances in the 
\abox, the abstract graph is always the same.
Looking at Table~\ref{tab:Empirical Results}, we can see that the number of states (in the Planning Graph and inconsistent ones) 
is greatly reduced with respect to FP.%, depicting a ``linear'' growth with the increase of the number of instances.

The abstract graph may show plans that cannot be found in the instantiated planning graph.
As an example of this, consider the following~$\abox$:\\ 
\idt$\{\cls{TechnicalDoc}(\cons{d001}),\cls{UrgentDoc}(\cons{d001}),\cls{Manager}(\cons{e001}),\cls{Administrative}(\cons{e003})\}$\\
which belongs to the set~$\concrete{S7}$ and in which we can perform the action $\acname{setAdmDoc}$.
This would lead, to the inconsistent $\abox$ where $\cons{d001}$ belongs both to $\cls{TechnicalDoc}$ and to $\cls{AdministrativeDoc}$,
which are disjoint concepts. Thus instantiated plans starting in {\em S7} cannot lead to the goal state.

%------------------

\begin{table}[h]
\small
\centering
\begin{tabular}{|c|c|c|c|c|c|c|c|c|c|c|c|c|}
\cline{1-3} \cline{5-8} \cline{10-13}
\multicolumn{3}{|c|}{\textbf{Instances}}    &  & \multicolumn{4}{|c|}{\textbf{FP Algorithm}}                                                  &           & \multicolumn{4}{|c|}{\textbf{ABP+FPI Algorithm}}                                               \\ \hline
$\cls{Mng}$ & $\cls{Emp}$ & $\cls{TechDoc}$ &  & \textit{$\vert P_{FP} \vert$} & \textit{$\vert V_{FP} \vert$} & \textit{Inc} & \textit{Time} & \textit{} & \textit{$\vert P_{FPI} \vert$} & \textit{$\vert V_{FPI} \vert$} & \textit{Inc} & \textit{Time} \\ \hline
1           & 1           & 1               &  & 3                             & 17                            & 13           & 0.06          &           & 3                              & 7                              & 3            & 0.07          \\ \hline
1           & 1           & 2               &  & 9                             & 38                            & 29           & 0.48          &           & 5                              & 10                             & 4            & 0.30          \\ \hline
1           & 1           & 3               &  & 25                            & 87                            & 66           & 0.28          &           & 7                              & 13                             & 5            & 0.10          \\ \hline
1           & 2           & 2               &  & 50                            & 154                           & 116          & 0.71          &           & 10                             & 15                             & 4            & 0.15          \\ \hline
2           & 2           & 2               &  & 80                            & 172                           & 134          & 1.35          &           & 16                             & 16                             & 5            & 0.22          \\ \hline
2           & 2           & 3               &  & 270                           & 413                           & 291          & 3.42          &           & 22                             & 21                             & 6            & 0.18          \\ \hline
2           & 3           & 3               &  & 816                           & 1802                          & 1290         & 33.16         &           & 33                             & 28                             & 6            & 0.24          \\ \hline
\vdots           & \vdots           & \vdots               &  & \vdots                           & \vdots                          & \vdots         & \vdots         &           & \vdots                             & \vdots                             & \vdots            & \vdots          \\ \hline
20           & 20           & 20               &  & -                           & -                          & -         & $\infty$         &           & 8800                             & 862                             & 41            & 197.40          \\ \hline
\end{tabular}
\caption{Empirical results, timing given in seconds, $\infty$ means more than 200 seconds.}
\label{tab:Empirical Results}
\vspace{-6mm}
\end{table}

Table \ref{tab:Empirical Results} summarizes the experiments over different \abox-es, 
where we change the number of instances (shown in the column $\mathbf{Instances}$).
For both algorithms we indicate: (i)~the size $\vert P \vert$ of the produced planning graph $P$ 
(we do not consider the intermediate Abstract Planning Graph), (ii)~the number $\vert V \vert$ of visited states 
during the creation of $P$ (again, we do not consider the abstract states computed in the ABP phase), 
(iii)~the number $Inc$ of discarded inconsistent states, and (iv)~the computation time (measured in seconds)
obtained as an average of 10 runs over the same example.
The timings are note very high because the code is just a prototype, but they can give an idea of the reduction of the state-space.

The results are promising because the \emph{ABP+FPI Algorithm} performs better that the standard \emph{Forward Algorithm}.
In particular, the number $\vert P \vert$ of edges in the FP algorithm and the number of inconsistent states grows quickly with 
the increasing number of instances. Already 2 managers, 3 employees and 3 technical documents produce a plan with 816 edges,
discarding 1,245 inconsistent states.
Such a difference between the two algorithms, can be explained by the large number of redundant plans found by FP
(as discussed in Examples \ref{ex:planning,forward algorithm} and \ref{ex:planning, backward algorithm}).

%The ABP+FPI, even if showing better performances over the FP, it's still slow for a real-case scenario:
%we ran an experiment using 20 instances for each class, and we obtained a graph composed 8,800 edges (although only 41 states were discarded as inconsistent), computed in 197.4 seconds (more than 3 minutes).

% -----------------

%\newpage
\vspace{-4mm}
\section{Conclusions and Future Work}
\label{sec:Conclusions}

%\vtodo{
%The approach taken by the authors seems to employ a backward search
%inside the planner using a reasoner for OWL Web ontology language
%called Pellet. It would be great if the authors would have
%demonstrated the strength of their knowledge representation techniques
%against well known AI planning systems such as STRIPS­planning. I
%expect using such a very heavy weight reasoner will have a major
%impact compared to using tools such as SAT solvers to solve similar
%problems.
%}
%
%\vtodo{Per rispondere ai dubbi riguardo all'utilità di usare DL e un full-reasoner, magari dire che questo è solo un primo passo per formalizzare come il planning può essere fatto con DL.
%Il passo successivo è ampliare l'utilizzo del reasoning offerto da DL per rendere il planning più raffinato a fronte di una semplificazione del costo di rappresentazione del mondo. In pratica ho una rappresentazione più ricca tramite i costrutti di DL, non ho bisogno di specificare tutto in quanto tramite reasoning sono in grado di ricavare informazioni nuove, nuova conoscenza, ed usarla per il planning.}

In this paper we have presented some preliminary work on a technique for reducing the state space in planning problems by exploiting a symbolic representation of states and reasoning techniques provided by Description Logics.
Although we have chosen to adopt the DL-Lite framework and Pellet as the reasoner of our implementation, we developed the Backward State Space Reduction technique to be as independent as possible from the actual reasoning mechanism of the underlying logical representation of knowledge.
The implementation of the ABP+FPI algorithm, compared to a standard Forward Planning algorithm, shows promising results both in terms of the time taken for finding the entire set of plans and in terms of the actual number of explored states.

There is a number of directions in which this work can be extended.
Currently, we are working on proving the relative completeness of our ABP+FPI algorithm
w.r.t.~the Forward Planning algorithm, when we restrict to non-redundant plans.
In the short term, we want to study the extension of actions that allows also to {\em remove}
\abox{} assertions. Afterward, we plan to study the extension of our technique to the more
general Datalog$^\pm$ family of languages as well as to allow the {\em creation} of new
individuals as an effect of actions, thus introducing a possibly infinite planning space.
%
%
%\vtodo{
%(1)~action guard defined as an UCQ + ECQ,
%(2)improve information extraction from KB: semantic connections, maybe avoid using SJ axioms
%}

\bigskip
\textbf{Acknowledgments.}
The research presented in this paper has been partially funded by the EU 
project ASCENS (nr.257414) %and QUANTICOL (nr.600708), 
and by the Italian MIUR PRIN project CINA (2010LHT4KM).

% -----------------

%\nocite{*}
\bibliography{dl}%,generic}
\bibliographystyle{eptcs}

% -----------------

%\newpage
\section*{Appendix}
\label{sec:Appendix}

\noindent The complete specification of the Case Study.

\noindent The $\tbox$ is the following:
\begin{multicols}{2}
\noindent
$\cls{Document} \sqsubseteq \neg \cls{Employee}$ \\
$\cls{Document} \sqsubseteq \neg \cls{DocumentState}$ \\
$\cls{DocumentState} \sqsubseteq \neg \cls{Employee}$ \\
$\cls{Technician} \sqsubseteq \cls{Employee}$ \\
$\cls{Administrative} \sqsubseteq \cls{Employee}$ \\
$\cls{Manager} \sqsubseteq \cls{Employee}$ \\
$\cls{Technician} \sqsubseteq \neg \cls{Administrative}$ \\
$\cls{Technician} \sqsubseteq \neg \cls{Manager}$ \\
$\cls{Administrative} \sqsubseteq \neg \cls{Manager}$ \\
$\cls{TechnicalDoc} \sqsubseteq \cls{Document}$ \\
$\cls{AdministrativeDoc} \sqsubseteq \cls{Document}$ \\
$\cls{UrgentDoc} \sqsubseteq \cls{Document}$ \\
$\cls{TechnicalDoc} \sqsubseteq \neg \cls{AdministrativeDoc}$ \\
$\cls{Technician} \sqsubseteq \exists \rol{canManage}.\cls{TechnicalDoc}$ \\
$\cls{Administrative} \sqsubseteq \exists \rol{canManage}.\cls{AdministrativeDoc}$ \\
$\cls{Document} \sqsubseteq \exists \rol{canManage}^-$ \\
$\exists \rol{canManage}^- \sqsubseteq \cls{Document}$ \\
$\exists \rol{assignedTo} \sqsubseteq \cls{Document}$ \\
$\exists \rol{assignedTo}^- \sqsubseteq \cls{Employee}$ \\
$\funct{\rol{assignedTo}}$ \\
$\exists \rol{hasStatus} \sqsubseteq \cls{Document}$ \\
$\exists \rol{hasStatus}^- \sqsubseteq \cls{DocumentState}$
\end{multicols}

\smallskip

\noindent The SJ axioms are: \\
$\cls{Technician}(x) \wedge \cls{TechnicalDoc}(y) \rightarrow \rol{canManage}(x,y)$ \\
$\cls{Administrative}(x) \wedge \cls{AdministrativeDoc}(y) \rightarrow \rol{canManage}(x,y)$

\smallskip

\noindent The $\abox$ is the following:
\begin{multicols}{3}
\noindent
$\cls{Manager}(\cons{e001})$ \\
$\cls{Technician}(\cons{e002})$ \\
$\cls{Administrative}(\cons{e003})$ \\
$\cls{TechnicalDoc}(\cons{d001})$ \\
$\cls{UrgentDoc}(\cons{d001})$ \\
$\cls{DocumentState}(\cons{reviewed})$
\end{multicols}

\smallskip

\noindent The available set $\Gamma$ of actions is: \\
\action{appoint}{x,y,z}{$\cls{Manager}(x) \wedge \rol{canManage}(y,z)$}{$\rol{assignedTo}(z,y)$} \\
\action{review}{x,y}{$\rol{assignedTo}(x,y)$}{$\rol{hasStatus}(x,\cons{reviewed})$} \\
\action{setAdmDoc}{x,y}{$\cls{Manager}(x) \wedge \cls{Document}(y)$}{$\cls{AdministrativeDoc}(y)$} \\
\action{setTechnician}{x,y}{$\cls{Manager}(x) \wedge \cls{Employee}(y)$}{$\cls{Technician}(y)$}

\smallskip

\noindent The goal is:\\
$goal: \rol{hasStatus}(x, \cons{reviewed}) \wedge \cls{UrgentDoc}(x)$

\end{document}